\documentclass[10pt,twocolumn,letterpaper]{article}

\usepackage{cvpr}              % To produce the CAMERA-READY version
\definecolor{cvprblue}{rgb}{0.21,0.49,0.74}
\usepackage[pagebackref,breaklinks,colorlinks,allcolors=cvprblue]{hyperref}
\usepackage[accsupp]{axessibility}
\usepackage{algorithm}
\usepackage{algorithmic}
\usepackage{xcolor}
\usepackage{pifont}
\usepackage{tabularx}
\usepackage[export]{adjustbox}
\usepackage{multirow}
\definecolor{darkergreen}{RGB}{21, 152, 56}
\definecolor{red2}{RGB}{252, 54, 65}
\newcommand{\yesmark}{\textcolor{darkergreen}{\ding{52}}}
\newcommand{\nomark}{\textcolor{red2}{\ding{56}}}

\def\paperID{1147} % *** Enter the Paper ID here
\def\confName{CVPR}
\def\confYear{2025}

\title{Classifier-guided CLIP Distillation for Unsupervised Multi-label Classification}

\author{Dongseob Kim\\
Samsung Electronics, Republic of Korea\\
{\tt\small kou.kim@samsung.com}
\and
Hyunjung Shim\\
KAIST, Republic of Korea\\
{\tt\small kateshim@kaist.ac.kr} 
}

\begin{document}
\maketitle
\begin{abstract}
Multi-label classification is crucial for comprehensive image understanding, yet acquiring accurate annotations is challenging and costly. To address this, a recent study suggests exploiting unsupervised multi-label classification leveraging CLIP, a powerful vision-language model. Despite CLIP's proficiency, it suffers from view-dependent predictions and inherent bias, limiting its effectiveness. We propose a novel method that addresses these issues by leveraging multiple views near target objects, guided by Class Activation Mapping (CAM) of the classifier, and debiasing pseudo-labels derived from CLIP predictions. Our Classifier-guided CLIP Distillation (CCD) enables selecting multiple local views without extra labels and debiasing predictions to enhance classification performance. Experimental results validate our method's superiority over existing techniques across diverse datasets. The code is available at https://github.com/k0u-id/CCD.
\end{abstract}    
\section{Introduction}
\label{sec:intro}

Multi-label classification plays a pivotal role in identifying multiple classes within an image, serving as a foundational component in image understanding~\cite{tsoumakas2007multi, misra2016seeing, liu2015optimality, sun2017revisiting, mahajan2018exploring}. Its applications are widespread, including recommendation systems, surveillance, and image retrieval~\cite{wang2009multi}. However, acquiring multi-label annotations poses challenges due to the difficulty and significant costs involved in meticulously annotating all relevant objects within an image.

To solve this problem, various label-efficient techniques have been proposed. A partial label setting~\cite{chen2022structured, kim2023bridging, durand2019learning, huynh2020interactive, pu2022semantic, 10168295} annotates only a portion of the classes appearing in an image and considers the rest as unknown labels. In the extreme, single positive label settings~\cite{cole2021multi, kim2022large, zhou2022acknowledging, Verelst_2023_WACV, song2025ignore} provide only one positive label per image, with all other labels assumed unknown. However, these methods still incur labeling costs. Besides, current studies simulate partial and single positive settings by randomly omitting labels from a fully labeled dataset, which may not accurately reflect the actual labeling practices. For example, humans tend to prioritize labeling trivial objects first.

To overcome these limitations, we focus on an unsupervised multi-label classification problem leveraging a vision-language model. One such model is CLIP~\cite{radford2021learning}, trained on an extensive dataset of 400M \{image, text\} pairs from the web. As a result, this model excels at recognizing diverse visual attributes and object categories in a zero-shot manner. Most recently, CDUL~\cite{abdelfattah2023cdul} applied CLIP for unsupervised multi-label classification by using it as a pseudo-labeling function. CDUL generates pseudo-labels for both global and local views to help identify less prominent objects, training a multi-label classifier with each view-label pair. However, CDUL overlooks the limitations of CLIP, which hinders its prediction performance. 
\begin{figure}[t]
\begin{center}
\includegraphics[width=\linewidth]{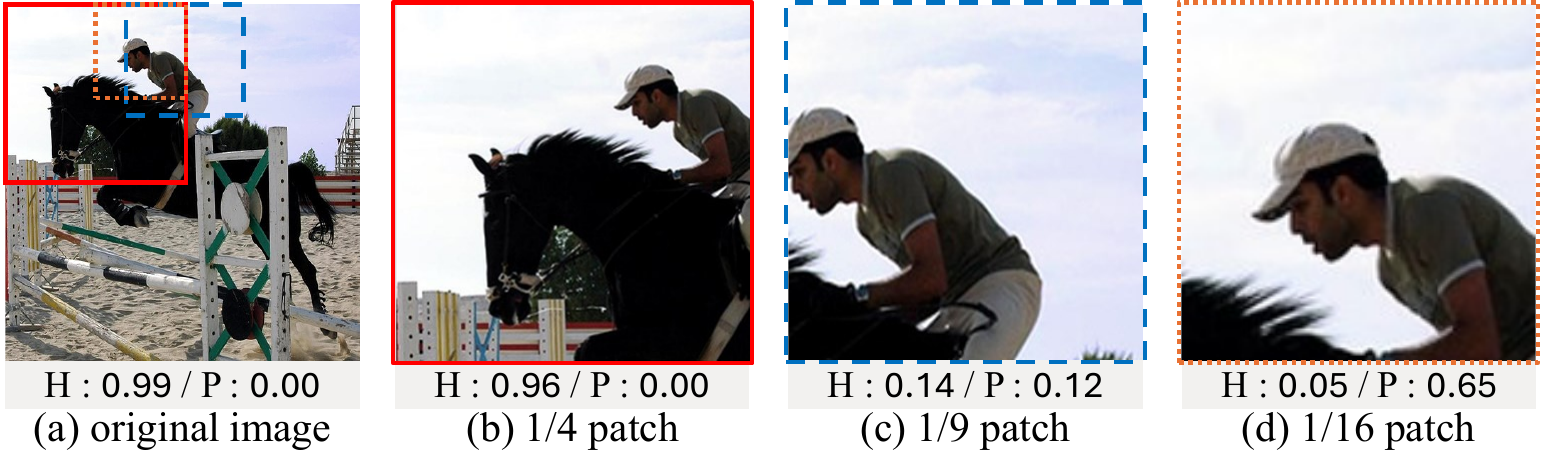}
\end{center}
\caption{CLIP prediction probability of \textit{horse} (H) and \textit{person} (P) corresponding to different input patches from single image: (a) The original image and its CLIP prediction. (b) The $1/4$ sized patch and its CLIP prediction. (c) The $1/9$ sized patch and its CLIP prediction. (d) The $1/16$ sized patch and its CLIP prediction. The same colored box indicates each patch cropped from the corresponding box.}%We can observe the slight translation and zoom lead to significantly different CLIP predictions.}
\label{fig:toy}
\end{figure}

In this paper, we identified two reliability issues in CLIP, addressing them to fully harness CLIP's potential. Firstly, we demonstrate that CLIP's predictions are highly view-dependent, considerably impacting classification performance. In a toy experiment, we observe that changing the crop position and size of an image leads to notable changes in CLIP's predictions. For example, in Fig.~\ref{fig:toy}, when the cropped area is slightly shifted or increased, the probabilities for the horse and person classes vary significantly. Based on these key observations, we use multiple local views near the target object to mitigate the uncertainty of CLIP's pseudo-labels caused by view changes. Specifically, we acquire various local views extracted by the bounding box using the Class Activation Mapping (CAM)~\cite{zhou2016learning} method, where the CAM is computed from the trained classifier itself. Additionally, our method can draw more local views for hard samples whose prediction exceeds a certain threshold across multiple classes.

\begin{figure}[t]
\begin{center}
\includegraphics[width=\linewidth]{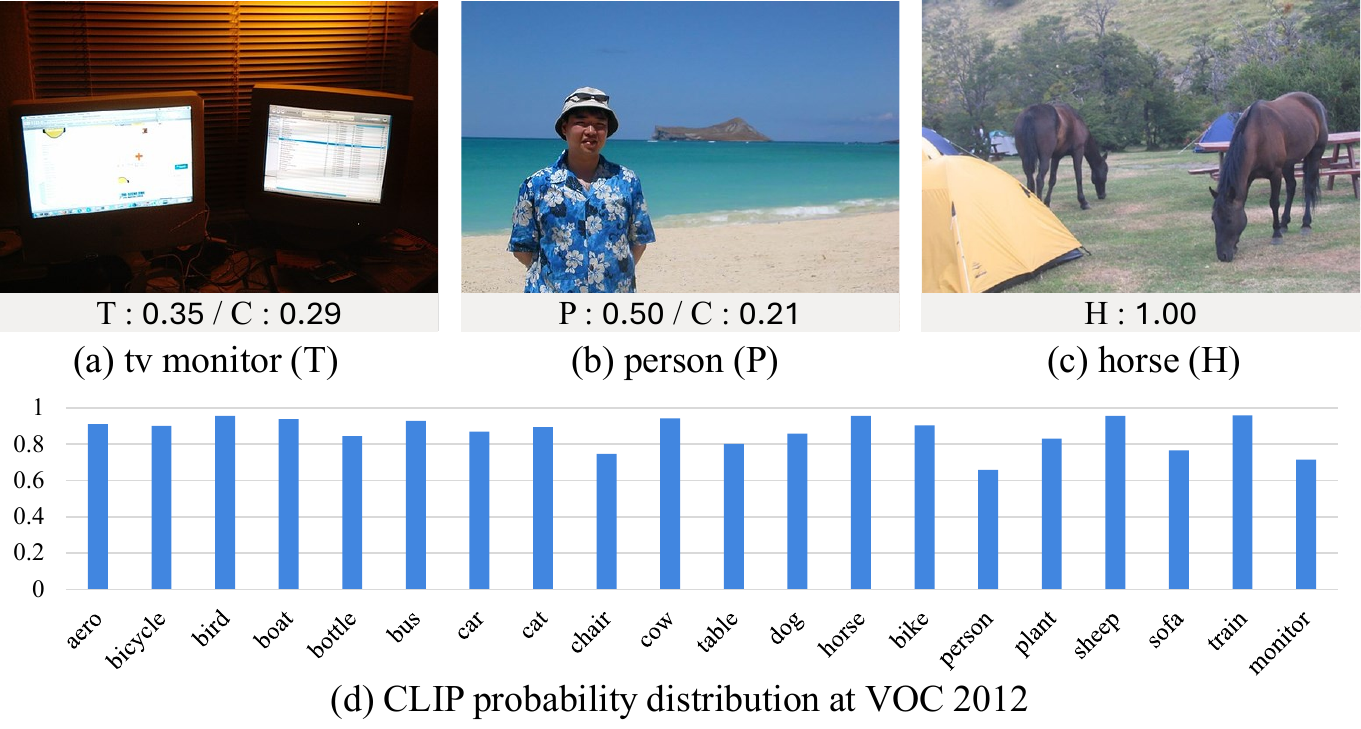}
\end{center}
\caption{Sample images showcasing CLIP bias. C indicates the probability of \textit{chair}. (a) Top 1 probability of ``tv monitor'' image is 35\%. (b) Top 1 probability of a ``person'' image is 50\%. (c) Top 1 probability of a ``horse'' image is 100\%. (d) The mean class-wise probability of PASCAL VOC 2012. We can observe the class-wise prediction bias of CLIP from these results.}
\label{fig:bias}
\end{figure}

Another notable issue with CLIP is its inherent prediction bias, which tends to reduce the predicted probability of specific classes. Due to polysemy~\cite{radford2021learning}, the words representing each class can be ambiguous in text embedding space. As illustrated in Fig.~\ref{fig:bias} (a) and (b), despite the presence of a single class (\textit{tv monitor} and \textit{person}, respectively), CLIP assigns considerably low top-1 probabilities of around 35\% and 50\%. Conversely, objects like horses exhibit notably higher probabilities, as depicted in Fig.~\ref{fig:bias} (c). Analyzing the average CLIP top-1 probabilities across classes, as shown in Fig.~\ref{fig:bias} (d), further highlights this bias, indicating a lower prediction for certain classes compared to others. The biased prediction is evident across various datasets such as MS COCO and NUSWIDE\footnote{CLIP probability distribution of other datasets are in the supplementary material.}. Since the proposed method relies on pseudo-labels generated by CLIP, its bias naturally transfers to the classifier. Consequently, the classifier shows a similar prediction bias and performance degradation. To mitigate this issue, we adjust all pseudo-labels by applying the inverse transform of the bias. While this adjustment can amplify unwanted noise in pseudo-labels, we apply consistency loss to handle this noise after debiasing. Our adjustment helps debiasing the classifier, thereby enhancing its generalization performance. As a result, our proposed Classifier-guided CLIP Distillation (CCD) achieves state-of-the-art performances on the PASCAL VOC 2012, PASCAL VOC 2007, COCO, and NUSWIDE benchmarks. 

\section{Related Work}

\subsection{Label efficient multi-label classification.}

Unlike multi-class classification, where a single class is identified in an image, multi-label classification requires accurately identifying all classes present in an image. Labeling every class in an image is significantly more challenging and resource-intensive than identifying just one class. To address this issue, a partial label setting has been proposed~\cite{chen2022structured, kim2023bridging, durand2019learning, huynh2020interactive, pu2022semantic, 10168295}, which utilizes only a subset of labels associated with each image.

Early approaches assumed any unannotated labels (\ie ``unknown labels'') as negative labels~\cite{5995734, Sun_Zhang_Zhou_2010, wang2014binary}. These methods involved correcting these labels during the learning process. However, approaches that relied on probabilistic models~\cite{deng2014scalable, wu2015ml, kapoor2012multilabel, vasisht2014active} for label correction were dependent on optimization techniques, resulting in limited scalability. Subsequently, alternative methods have been proposed, such as leveraging label-to-label similarity~\cite{huynh2020interactive}, cross-image semantic correlation~\cite{chen2022structured}, or category-specific representation~\cite{pu2022semantic} to refine labels. Some approaches suggest using soft-pseudo-labels instead of unknown labels~\cite{szegedy2016rethinking, mac2019presence, durand2019learning}.

Research has extended to a more extreme scenario: single positive label setting~\cite{cole2021multi, kim2022large, zhou2022acknowledging, Verelst_2023_WACV}, where only a single positive label is provided per image. These methods build upon those proposed in the partial label setting and adapt them to function effectively in more challenging contexts.

More recently, CDUL~\cite{abdelfattah2023cdul} has addressed the unsupervised multi-label classification problem by utilizing CLIP. In a similar vein, our method tackles the unsupervised multi-label classification problem using CLIP while additionally providing local area information with CAM and addressing bias in CLIP predictions.

\subsection{Vision-language pretraining.}
Vision-language pretraining (VLP) has emerged as a pivotal approach for bridging the gap between visual and textual modalities~\cite{miech2020end, singh2022flava, jia2021scaling}. Among the pioneering VLPs, CLIP~\cite{radford2021learning} stands out as the most commonly utilized framework. CLIP was trained on large \{image, text\} pairs sourced from the Internet. Recently, several studies have adopted VLP to provide rich visual-semantic knowledge for downstream image tasks, such as detection~\cite{gu2021open, zareian2021open}, segmentation~\cite{kim2023weakly, rao2022denseclip, zhou2022extract}, human-object interation~\cite{khandelwal2022simple, liao2022gen}, and generation~\cite{crowson2022vqgan, tevet2022motionclip, wang2022clip}. 

In multi-label classification, various methods have emerged to derive pseudo-labels from CLIP predictions~\cite{abdelfattah2023cdul, xing2023vision, hu2023dualcoop++}. Among these, CDUL~\cite{abdelfattah2023cdul} integrates local inferences with global inference, successfully expanding the utilization of the VLP to multi-label classification. However, CDUL fails to tackle two limitations inherent in CLIP: sensitivity to input view selection and inherent prediction bias. Consequently, the pseudo-labels generated by CDUL exhibit noise and bias. In contrast, our approach leverages a fine-tuned classifier to suggest local views and addresses CLIP bias through calibration. 
\section{Method}

\subsection{Generating initial pseudo-labels}
CLIP, a powerful vision-language model, has been trained to align images with corresponding textual descriptions. Leveraging the powerful zero-shot capability of CLIP, we initially generate pseudo-labels similar to CDUL. For each image in the dataset, we extract its image embedding using the CLIP image encoder. Simultaneously, we obtain the text embedding by inputting the class name associated with each dataset and a fixed prompt (\eg ``a photo of the [class name]'') into the text encoder. Next, we compute the similarity between the image embedding and the class-specific text embedding. This similarity score is then translated to compute softmax probabilities. These probabilities serve as the initial pseudo-labels. Mathematically, the similarity score for each class can be expressed as:

\begin{figure}[t]
\begin{center}
\includegraphics[width=\linewidth]{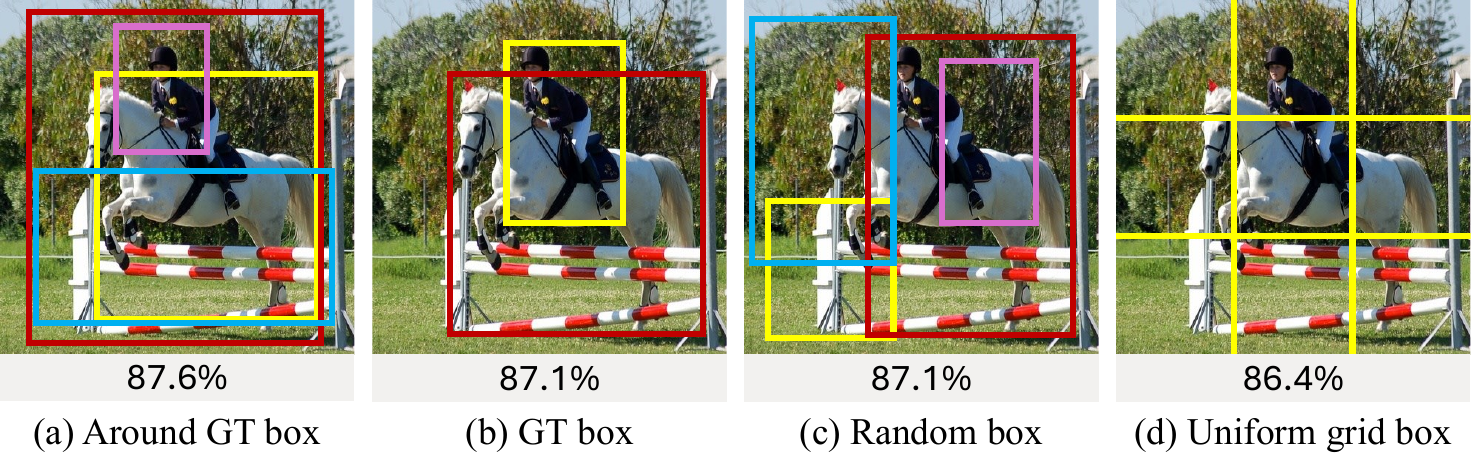}
\end{center}
\caption{The proof-of-concept study of local view selection method. We train the classifier with pseudo-label generated from four different local views: (a) Around GT boxes, (b) GT boxes, (c) Random boxes, (d) Uniform grid boxes. The numbers below each sample are the performance (mAP) of the classifier trained with the corresponding pseudo-label for the entire training set. The classifier trained with local view around GT boxes achieved 1.3\%p higher performance compared to classifier trained with uniform-grid local view.}
\label{fig:concept}
\end{figure}

\begin{equation}
    s_{c} = \frac{f^{\top}w_{c}}{\left \| f \right \|\cdot\left \| w_{c} \right \| }, \ \  1\leq c \leq C 
\end{equation}

\noindent where $f$ is a CLIP image embedding, $w_c$ is a class-specific CLIP text embedding and $c$ indicates a class index. With class-wise similarity, we can calculate a class-wise softmax probability $p_c$ of each image as follows:

\begin{equation}
    p_{c} = \frac{\mathrm{exp}(s_{c}/\tau )}{\sum^{C}_{c=1}\mathrm{exp}(s_c/\tau)},
\end{equation}

\noindent where $\tau$ is a CLIP temperature parameter. We concatenate each class-wise softmax probability to generate initial pseudo-labels. Finally, the initial pseudo-label for each image is defined as:

\begin{equation}
     l^{initial} = \left \{p_1^{global},p_2^{global},\cdots,p_c^{global} \right \}. 
\end{equation}

Following the acquisition of softmax probabilities, the initial pseudo-labels are used for the warm-up training of the classifier. Since the softmax calculation inherently normalizes the probabilities to ensure a sum of 1, it does not fit the multi-label classification problem where multiple classes have the same high probability (\eg 1). Nevertheless, after the classifier undergoes warm-up training, it can predict each class independently, leading to recognizing multiple classes present in an image, albeit with limited accuracy. To address this performance limitation, in the subsequent section, we introduce a method to construct \{training data, pseudo-label\} pairs tailored for the multi-label classification problem, leveraging a warm-up trained classifier.

\subsection{Updating classifier-guided pseudo-label}
Prior research~\cite{abdelfattah2023cdul} has shown that CLIP's inference results on the entire images may be suboptimal for multi-label classification. CLIP tends to assign high probabilities primarily to prominent objects in the input image. As a result, it does not effectively distribute probabilities across all existing objects. CDUL addresses this limitation by dividing the image into uniform patches in a grid fashion. Despite its simplicity, employing a uniform patch-based approach for local view acquisition may lead to the inference of background batch, where the patch does not contain any class object. More importantly, treating both simple images (\ie those containing a single object) and complex images (\ie those containing multiple objects) equally with the same number of local views is clearly inefficient in optimizing the resource (\ie CLIP inference).

\begin{figure*}[t]
\begin{center}
\includegraphics[width=\textwidth]{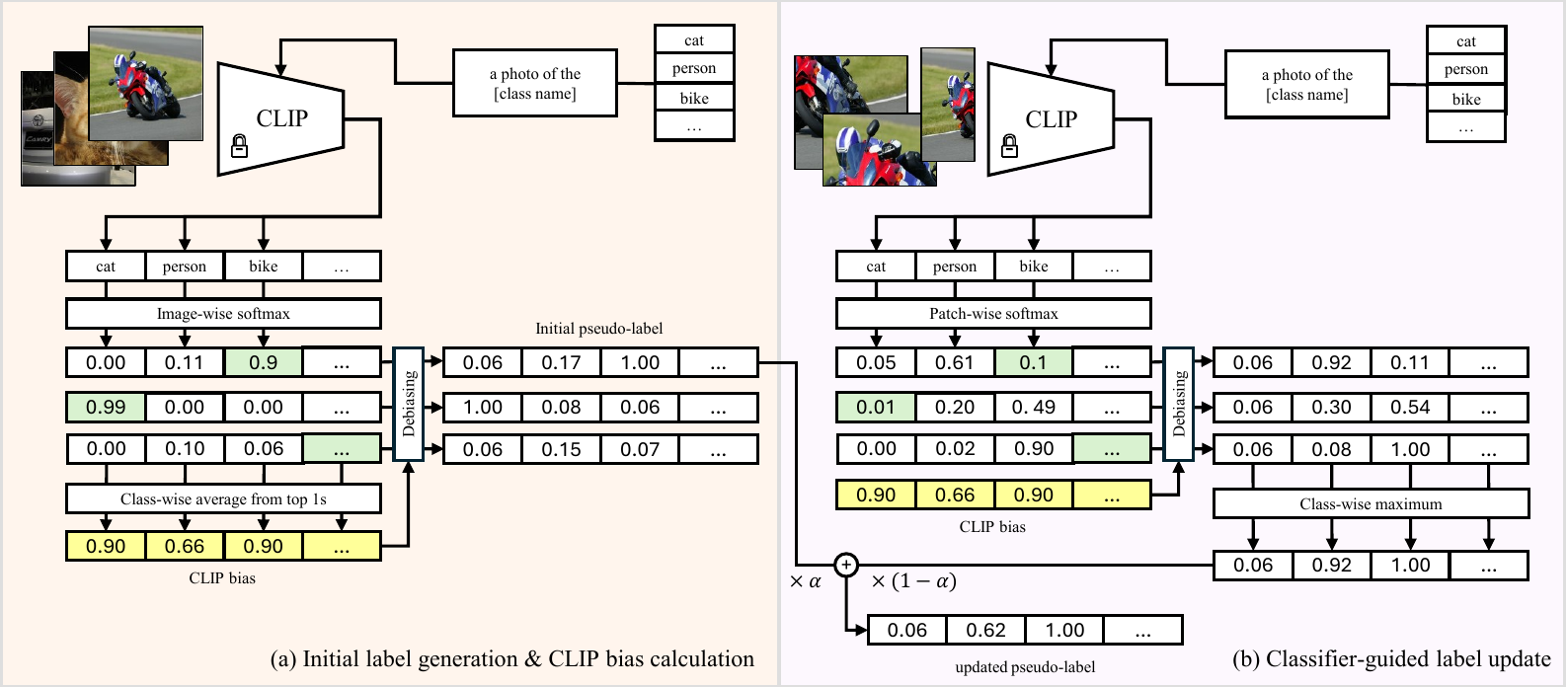}
\end{center}
\caption{The overview of label preparation. (a) We calculate the cosine similarity between text embeddings and image embeddings. The softmax probability of these similarities is the CLIP prediction of each image. The top 1 probabilities are highlighted in \textit{green boxes}. By class-wise averaging the top-1 probabilities, the CLIP bias is derived (\textit{yellow box}). Pseudo-labels are then generated by debiasing CLIP predictions. (b) For label updating, classes above a threshold are selected from the classifier output, and local views corresponding to these classes are extracted. The process for acquiring local labels for each patch mirrors the initial label acquisition. The final pseudo-label is obtained through a weighted sum of initial pseudo-labels and local labels.}
\label{fig:data_prepare}
\end{figure*}

\noindent \textbf{Proof-of-concept study.} We perform a proof-of-concept study to evaluate the impact of using object-specific locations in local view proposals. Our hypothesis is that incorporating class-related object locations enhances performance. We extract local views in four different scenarios, using ground truth bounding boxes (\ie, the bounding box that tightly covers the target object): (1) random cropping around the bounding box, (2) using only the bounding box itself\footnote{For images with fewer than nine boxes, we additionally infer their resized versions.}, (3) random cropping without guidance from the bounding box, and (4) uniform grid-based cropping. Each scenario produces nine patches per image. Subsequently, we infer pseudo-labels for these patches and train with each \{patch, pseudo-label\} pair. As shown in Fig.~\ref{fig:concept}, our results indicate that the first scenario--(a) random cropping around the bounding box--achieves the best performance. 

Our proof-of-concept study reveals two critical insights: (1) the local view generation policy significantly influences CLIP's predictions, and (2) framing the view around the target object greatly improves classification accuracy. 

\noindent \textbf{CAM-based view-selection.} Motivated by the above two findings, we decided to approximate the first scenario in our approach. We apply Class Activation Mapping (CAM) with the trained classifier to identify vital regions within the image. This is called a CAM-based view selection method. This method generates \{training image, pseudo-label\} pairs by focusing on the object periphery, without relying on any ground-truth annotations.

By employing the trained classifier, CAM visualizes activations that most significantly influence class prediction. CAM calculation is specifically based on the weights assigned to each channel of the feature map, which influence the classification score. This can be represented by the following formula:

\begin{equation}
    G_c(x) = \sum_{i=1}^{Q}\mathbf{w}^{c\top}_{i}\cdot g_i(x), 
\end{equation}

\noindent where $x$ is input image, $g$ is CNN network, $g(x)$ is the feature map of the CNN, $\mathbf{w}^{c}$ is class-wise weight of classifier, and $Q$ indicates the number of channels in the feature map.

We first normalize the activation map to a range between 0 and 1, then apply a threshold to identify highly activated regions. Next, we generate candidate bounding boxes with slight perturbation offsets. These bounding boxes define local views of the image. The regions highlighted by CAM act as reasonable approximations of the object's location, allowing us to extract local views without additional labeling costs. These localized view images are fed into CLIP to obtain softmax probabilities for the patches. The probabilities of local views are computed using the same method as for the global view, and then generating local labels that are focused on each local region.

\noindent \textbf{The number of local views.} Additionally, we adjust the number of local views based on image complexity. For simpler images, the classifier typically provides confident, high-probability predictions. This translates to high logits for well-trained classes and near-zero for others. Conversely, complex images often produce uncertain predictions, characterized by higher logits distributed broadly across multiple classes. To account for this, we conduct additional local view sampling when the classifier output of a certain class surpasses a predefined threshold. Specifically, we crop the bounding box of any class exceeding this threshold value and obtain its local probability. Through our experiments, we empirically determined a fixed threshold that ensures appropriate average number of classes exceeding this threshold. Details on this ablation study can be found in the experiments section. This approach allocates fewer local views to simple images and more local views to complex ones, optimizing the local view assignment. 

To obtain a single local pseudo-label for each input image, we select the maximum value of class-wise probability across all local patches from the corresponding image. Denoting the probability of $i-$th patch for class $c$ as $p^{patch}_{i,c}$, we conduct patch-wise aggregation using the maximum value as follows:

\begin{equation}
    p_c^{local} = \max_{i=1,\cdots,N}p^{patch}_{i,c}, \ \  1\leq c \leq C 
\end{equation}

\noindent where $N$ represents the number of local patches corresponding to each image, dependent on the input image itself. We then aggregate the local class-wise probability for generating a local pseudo-label as follows:

\begin{equation}
     l^{local} = \left \{p_1^{local},p_2^{local},\cdots,p_c^{local} \right \}. 
\end{equation}

\noindent We update the pseudo-label by the weighted sum of initial and local pseudo-labels. The final pseudo-label is calculated as follows: 

\begin{equation}
    l^{final}=\alpha l^{initial}+(1-\alpha)l^{local},
    \label{eqn:weightsum}
\end{equation}

\noindent where $\alpha$ is a hyperparameter and we used 0.4. Through this label update, the pseudo-label efficiently incorporates information about less salient objects in multi-label classification. The updated pseudo-label is utilized throughout the remainder of the training process.

\subsection{Debiasing CLIP pseudo-label}
Earlier, we aimed to identify highly activated areas within images through Class Activation Mapping and to infer additional label information focused on these areas. However, a challenge arises when using pseudo-labels generated by CLIP due to its inherent prediction bias. Consequently, classifiers trained using these pseudo-labels may inadvertently inherit this bias. Previous work~\cite{algan2020label} reveals that while classifiers are robust at overcoming random noise, they cannot rectify consistent biases associated with specific classes. Therefore, we need to overcome the inherent prediction bias of CLIP to enhance the performance.

To understand this bias, given CLIP's focus on salient objects, we treat global CLIP prediction as a single-label classification task. In a single-label classification, the predicted probability distribution for each class aligns with the class-wise average of the highest probabilities from each prediction. Following this principle, we compute the average prediction probability for each class, defining it as CLIP's prediction bias. Specifically, we compute CLIP bias as follows: (1) Obtain initial pseudo-labels by running each image in the ``target training set'' through CLIP; (2) Record top-1 probabilities for samples below the entropy threshold to constrain the task to single-label classification; (3) Count each class occurrences only if it achieves the top 1 probability; (4) Normalize each class's softmax probability by its occurrence count, defining this normalized probability as CLIP bias. This process does not require any ground-truth annotation, solely relying on the result of CLIP prediction.

The bias obtained in this manner serves as a basis for calibrating CLIP-derived probabilities. Specifically, we calibrate by multiplying each class's softmax probability from CLIP by the inverse of the CLIP bias. This calibration process enables the classifier to learn a rectified probability distribution. Our debiasing technique leads to a noticeable improvement in the performance of classes previously challenged due to the inherent CLIP bias (see Tab.~\ref{tab:biasedclass}). Furthermore, we adjust excessively small probabilities to prevent the minimum probability value from approaching zero, aiding in learning the corrected distribution. In doing so, we guide the classifier as a whole to learn from this inverse distribution. This calibration is applied not only to the global probability but also to local inference, minimizing the influence of the CLIP bias on the classifier as much as possible.

\subsection{Overall training}

\begin{figure}[t]
\begin{center}
\includegraphics[width=\linewidth]{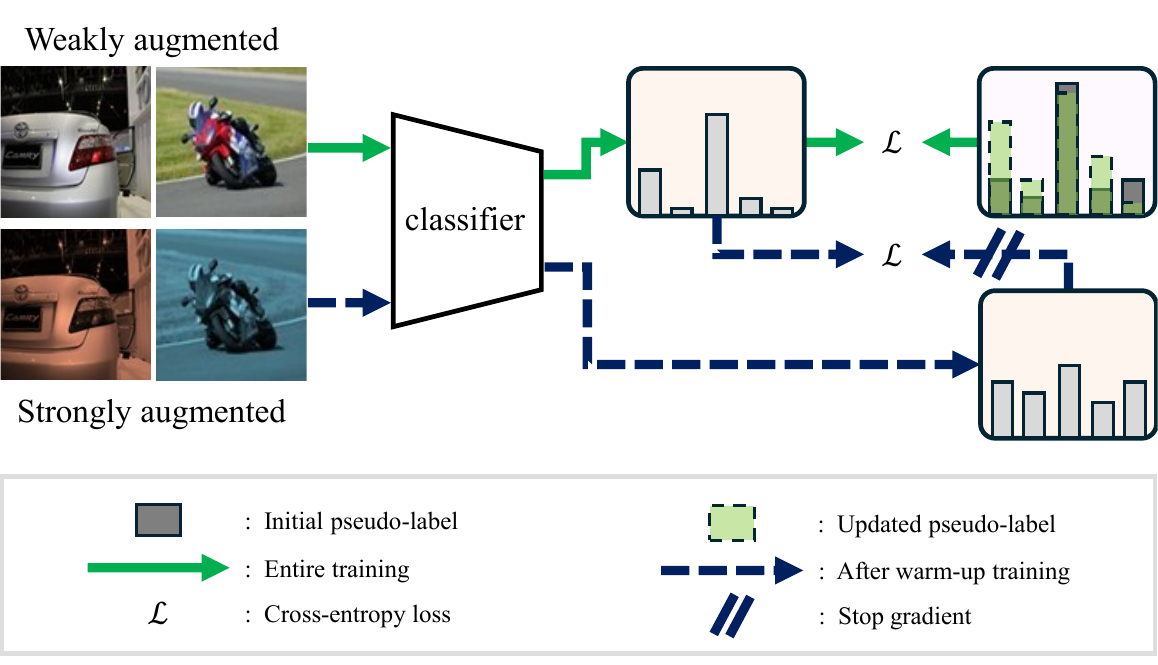}
\end{center}
\vspace{-5mm}
\caption{The training process of our method. We train the classifier using cross-entropy targeting initial pseudo-label during the warm-up phase which is illustrated as the \textit{green line}. After the classifier-guided label update, we train the classifier using the cross-entropy loss targeting updated pseudo-label and the cross-entropy loss between the logits of differently augmented inputs which are illustrated as the \textit{green line} and the \textit{blue dashed line}.}
\label{fig:process}
\end{figure}
\noindent \textbf{Consistency loss.} The proposed method tackles noisy data by utilizing pseudo-labels as target labels for all \{training data, pseudo-label\} pairs. Since pseudo-labels inherently carry noise and our debiasing process can potentially amplify this noise, we employ consistency loss between the data and its augmented data as additional critics. This approach is inspired by the common practice in a semi-supervised regime, improving feature extractors for handling noisy data. In our framework, we introduce consistency loss alongside the conventional binary cross-entropy loss during training. Consistency loss involves applying different augmentation intensities to the same image. Also, it passes each augmented version through the classifier to ensure consistent predictions. In Fig.~\ref{fig:process}, we employ weak augmentations, such as resizing and flipping, and strong augmentations, which comprise weak augmentation and additional color jittering.

\noindent \textbf{Training process.} Initially, the entire training dataset undergoes inference through CLIP, followed by calculating the CLIP bias. Subsequently, we derive a global label by debiasing the global probability, utilizing it as an initial pseudo-label in Fig.~\ref{fig:data_prepare} (a). The classifier is then warmed up with this pseudo-label as the target (the green line in Fig.~\ref{fig:process}). Following warm-up training, we update the pseudo-label with classifier-guided local inference. During this phase, the obtained local probabilities undergoes debiasing. Then they are combined to derive a local label, which then updates the existing pseudo-label as depicted in Fig.~\ref{fig:data_prepare} (b). The classifier is subsequently trained with the updated pseudo-label for the remaining epochs (the green line and the blue dashed line in Fig.~\ref{fig:process}). For whole trainig, our loss function combines two terms:

\begin{equation}
    \mathcal{L}_{total}= L_{ce}+ \beta L_{consist},
    \label{eqn:loss}
\end{equation}

\noindent where $\mathcal{L}_{ce}$ is cross-entropy loss and $\mathcal{L}_{consist}$ is consistency loss. $\beta$ is a hyperparameter and we fix it as 0 during warm-up phase and 1 after.% This way, both loss functions are weighted equally during training.
\section{Experiments}

\subsection{Experimental setup}

\noindent\textbf{Dataset \& evaluation Metric.}
We conducted evaluations on four distinct datasets. The PASCAL VOC 2012~\cite{Everingham10} dataset comprises 20 classes, with 5,717 images in the training set and 5,823 in the validation set. Similarly, the PASCAL VOC 2007 dataset also encompasses 20 classes, with a combined total of 5,011 training and validation images and 4,952 test images. MS COCO~\cite{lin2014microsoft} consists of 80 classes, encompassing 82,081 training images and 40,137 validation images. Additionally, the NUSWIDE~\cite{chua2009nus} dataset comprises 81 classes, with a training set of 150,000 images and a test set of 60,200 images. In cases where a separate test set was not available, the validation set was utilized for testing purposes. Mean Average Precision (mAP) was employed as the performance metric, with ground truth labels utilized solely for performance evaluation and not incorporated during the model training phase.

\begin{table*}[!t]
    \centering{
    \caption{Mean average precision mAP in (\%) for different multi-label classification methods under different supervision levels. Bold represents the best results among unsupervised methods. The proposed achieved state-of-the-art performance.}
    \label{tab:quantitative}
\begin{adjustbox}{width=\textwidth}
    \begin{tabular}%{|l|l|l|l|l|}
        {
    %   |
      >{}p{0.2\textwidth}|
      >{}p{0.16\textwidth}|
     >{}p{0.19\textwidth}|
      >{\centering}p{0.10\textwidth}|
     >{\centering}p{0.10\textwidth}|
      >{\centering}p{0.10\textwidth}|
      >{\centering\arraybackslash}p{0.10\textwidth}
    %   |
    }
\hline

        Supervision level & Annotation & Method   & VOC12 & VOC07 & COCO & NUS  \\ \hline \hline
                %%%%%%%%%%%%%%%%%%%%%%%%%%%%%%%%%%%%
        \multirow{2}{*}{Fully supervised}  & \multirow{2}{*}{Fully labeled}  
        %bce
        & BCE  &90.1 &91.3 &78.5 & 50.7 \\ \cline{3-7}
        %bce
        ~& ~& BCE-LS~\cite{cole2021multi} &91.6 &92.6 &79.4 &51.7  \\ \hline \hline
        %%%%%%%%%%%%%%%%%%%%%%%%%%%%%%%%%%%
         \multirow{5}{*}{Weakly supervised}  & \multirow{3}{*}{\begin{tabular}[c]{@{}l@{}}Partial labeled\\ (10\%)\end{tabular}} & SARB~\cite{pu2022semantic} & - &85.7 & 72.5 & -  \\ \cline{3-7}
        %AAAI (10% partial labels resnet-101)
        ~ & ~  &ASL~\cite{ridnik2021asymmetric} & - &82.9 & 69.7 & -  \\ \cline{3-7}
        ~ & ~  &Chen \textit{et al.}~\cite{chen2022structured} & - &81.5 & 68.1 & -  \\ \cline{2-7}    
        ~ & \multirow{2}{*}{\begin{tabular}[c]{@{}l@{}}Single positive\\ labeled\end{tabular}}  &  LL-R~\cite{kim2022large} & 89.7 & 90.6 & 72.6 & 47.4 \\ \cline{3-7}
        ~ & ~ &  G$^2$NetPL~\cite{abdelfattah2022g2netpl} & 89.5 &89.9  & 72.5 & 48.5 \\ \hline \hline

         \multirow{6}{*}{Unsupervised} & \multirow{6}{*}{\begin{tabular}[c]{@{}l@{}}Annotation\\ free\end{tabular}} & Naive AN~\cite{kundu2020exploiting}  & 85.5 &86.5 & 65.1 & 40.8 \\ \cline{3-7}
         %ANLS
        & ~ & Szegedy \textit{et al.}~\cite{szegedy2016rethinking} & 86.8 &87.9 & 65.5 & 41.3 \\ \cline{3-7}
        %WAN
         & ~ & Aodha \textit{et al.}~\cite{mac2019presence} & 84.2 &86.2 & 63.9 & 40.1 \\ \cline{3-7}
         %curriculum learning
         & ~ &   Durand \textit{et al.}~\cite{durand2019learning} & 81.3 &83.1 & 63.2 & 39.4 \\ \cline{3-7}
         %ROLE
         ~ & ~ & ROLE~\cite{cole2021multi} & 82.6 &84.6 & 67.1 & 43.2  \\ \cline{3-7}

         ~ & ~ & CDUL~\cite{abdelfattah2023cdul} & 88.6 & 89.0 & 69.2 & 44.0 \\ \cline{3-7}

         ~ & ~ & CCD (Ours) & \textbf{90.1} & \textbf{91.0} & \textbf{70.3} & \textbf{44.5} \\ \cline{1-7}
\hline
% \hline

    \end{tabular}
        \end{adjustbox}
    }
    %\vspace{-0.5 cm}
\end{table*}

\noindent\textbf{Implementation detail.}
For a fair comparison, we endeavored to set the hyperparameters for CLIP inference as closely as possible to those of the primary competitive method, CDUL. We acquired labels using the frozen CLIP ResNet50$\times$64, and employed the ImageNet pre-trained ResNet101 as the backbone network for our classifier. The entropy threshold for calculating CLIP bias and the classifier threshold for local view inference is both 0.5. It Our warm-up classifier is trained for two epochs. While the entire training process can proceed up to 10 epochs, we obtain our final model via early stopping to prevent overfitting. Specifically, we terminate training based on the gradient of mean Average Precision (mAP) of the \textit{training labels} throughout the learning process. Training was ceased when the gradient of mAP reached the first local minimum. We utilized a learning rate of $10^{-5}$ across all datasets and employed a batch size of 16. Undisclosed hyperparameters of CDUL were manually reproduced. For generating the initial pseudo-labels of global views, we maintained a fixed image resolution of 640$\times$640. The CLIP prompts consistently used the phrase ``a photo of the [class name]'' except for the NUSWIDE dataset. (For a fair comparison with the existing method, CDUL, we employ the phrase ``a photo of a [class name]'' to match the pseudo-label performance.) For acquiring the initial pseudo-labels of local views, we introduced an offset value of 80 pixels for each vertexes when generating bounding boxes (\ie a thresholded CAM-bounding box $\pm 80$ as our local views). Additionally, cropped images were resized to have a longer side of 640 pixels while preserving their aspect ratios to obtain CLIP probabilities.

\subsection{Comparison with state-of-the-art methods}

\noindent \textbf{Competitive methods.} Multi-label classification can be partitioned depending on the type of supervision: fully supervised, weakly supervised, and unsupervised approaches. Firstly, in a fully supervised setting, we choose (1) the Na\'ive method using only binary cross-entropy (BCE) loss and (2) the Na\'ive method with additional label smoothing (LS)~\cite{cole2021multi}. Secondly, we compared our method with recent weakly supervised methods, including partial labeled methods utilizing only 10\% of total labels such as (3) SARB~\cite{pu2022semantic}, (4) ASL~\cite{ridnik2021asymmetric}, (5) Chen et al.~\cite{chen2022structured}. For single positive label methods, we choose (6) LL-R~\cite{kim2022large}, (7) G$^2$NetPL~\cite{abdelfattah2022g2netpl}. Finally, in an unsupervised scenario, we obtained initial labels using CLIP and evaluated the performance of (8) Na\'ive AN~\cite{kundu2020exploiting}, (9) Szegedy et al.~\cite{szegedy2016rethinking}, (10) Aodha et al.~\cite{mac2019presence}, (11) Durand et al.~\cite{durand2019learning}, (12) ROLE~\cite{cole2021multi}, which leveraging CLIP generated initial labels. We conducted a comparative analysis with (13) CDUL~\cite{abdelfattah2023cdul}, the current state-of-the-art in an unsupervised multi-label classification task.

\noindent \textbf{Comparison with fully \& weakly methods.} Compared to fully supervised methods, our Classifier Guided CLIP Distillation (CCD) demonstrates comparable performance on the PASCAL VOC 2012 and PASCAL VOC 2007 datasets. In particular, achieving similar performance without any human annotation is impressive. CCD outperforms various weakly supervised methods on the PASCAL VOC 2012 and PASCAL VOC 2007 datasets, although our performances on MS COCO and NUSWIDE are relatively lower. This suggests that CCD is more effective on datasets with a smaller number of classes, such as PASCAL VOC 2012 and 2007, which consist of 20 classes. On the other hand, the performance gain is relatively small for complex datasets with many classes. It can be attributed to the simplistic prompt, ``a photo of the [class name],'' which might not accurately represent the target classes and co-occurrence issue, which bring out noisy local labels. The performance of initial labels of the training sets for PASCAL VOC 2012, MS COCO, and NUSWIDE is 85.3\%, 65.4\%, and 41.2\%, respectively. CCD consistently shows about a 3~4\%p performance gain over all datasets.

\begin{table}[!t] 
    \centering
    \caption{Ablation study of the proposed moduels. Mean average precision mAP in (\%) for different configuration uon the PASCAL VOC 2012 dataset. The best score is in bold.}
        \label{tab:abl_module}
    \begin{tabular}{
    >{\centering}p{0.2\linewidth}|
      >{\centering}p{0.2\linewidth}|
     >{\centering}p{0.2\linewidth}|
      >{\centering\arraybackslash}p{0.2\linewidth}}
\hline 
         \multirow{2}{*}{\begin{tabular}[c]{@{}c@{}}Label\\ update\end{tabular}} &\multirow{2}{*}{\begin{tabular}[c]{@{}c@{}}CLIP\\debias\end{tabular}} &\multirow{2}{*}{Consistency} & \multirow{2}{*}{\begin{tabular}[c]{@{}c@{}}mAP\\(VOC12)\end{tabular}} \\ 
          & & & \\ \hline \hline 
           \nomark & \nomark  & \nomark  &86.4  \\ \hline
           \yesmark &\nomark &\nomark & 88.7 \\ \hline
           \yesmark & \yesmark & \nomark& 89.4 \\ \hline
           \yesmark & \nomark& \yesmark & 88.8 \\ \hline
           \yesmark & \yesmark & \yesmark & \textbf{90.1} \\ \hline
%\hline
    \end{tabular}
\end{table}
%%%%%%%%%%%%%

\noindent \textbf{Comparison with unsupervised methods.} Compared to unsupervised methods, CCD exhibits superior performance across all datasets. Particularly noteworthy are the significant improvements of 1.5\%p, 2\%p, and 1.1\%p on PASCAL VOC 2012, PASCAL VOC 2007, and MS COCO, respectively, compared to the existing state-of-the-art methods. These results highlight the effectiveness of our method. However, in the case of the NUSWIDE dataset, the performance improvement is not as substantial. This can be attributed to the unique characteristics of the NUSWIDE dataset, where 20\% of the train/test images do not contain any class-related objects (\ie ``none images''). As a result, the initial pseudo-label acquisition process fails to form a high probability for specific classes in these images. Consequently, during local inference, a significant portion of inferences is assigned to ``none images,'' thereby reducing the quality of pseudo-labels of CCD.

\subsection{Ablation}

\begin{figure}[t]
\begin{center}
\includegraphics[width=\linewidth]{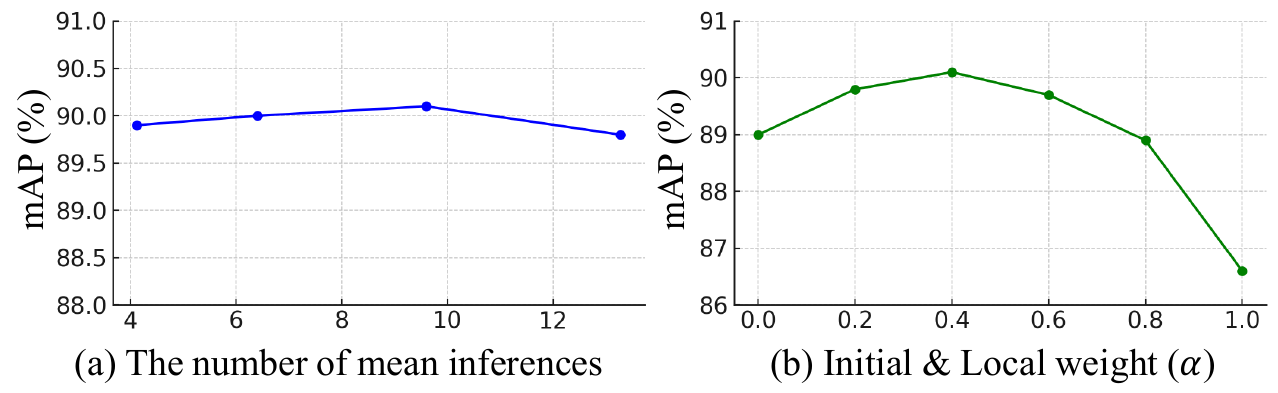}
\end{center}
\caption{Classification results (mAP) on PASCAL VOC 2012 depending on (a) the change of the number of mean inferences (b) the change of ratio of pseudo-label updating. }
\label{fig:plot}
\end{figure}

\noindent \textbf{Effects of each module.} We evaluate the effectiveness of each component of our method in Tab.~\ref{tab:abl_module}. The performance of the baseline model, the classifier trained with CLIP-generated global pseudo-labels, is 86.4\%. When we update the pseudo-label with classifier-guided local inference, it shows a significant improvement of 2.3\%p. This shows that our CAM-based local view selection method and adaptively allocating the number of CLIP inferences are effective. It already exceeds the performance of previous state-of-the-art methods. Debiasing CLIP also contributed 0.7\%p, indicating that our proposed debiasing method effectively relieves the CLIP bias. The consistency loss alone adds only 0.1\%p performance gain. However, when CLIP debiasing and consistency loss are combined, a substantial improvement of 1.4\%p is achieved. This result implies that pseudo-labels generated by compensating for CLIP bias may amplify both accurate information and noise. Since the consistency loss is effective for dealing with noisy data, it creates a synergy with CLIP debiasing.

\noindent \textbf{The number of inference.} We examine how model performance varies with different numbers of CLIP local inferences. We incrementally increase the average number of inferences to approximately 4, 6, 10, and 14. In Fig.~\ref{fig:plot} (a), the model achieves its best performance when the average number of inferences is around 10. Notably, increasing the number of inferences does not consistently enhance performance. Fewer inferences limit the allocation of local views to challenging samples, which hinders the ability to extract sufficient information from complex images, thereby negatively impacting negatively. In contrast, an excessive number of inferences results in the overallocation of local views, even for easy samples. It potentially introduces redundant inference for easy samples, making noisy pseudo-labels. In particular, when the average number of inferences is set to 10 or 14, the maximum number of local inferences remains capped at 20 (\ie the total number of classes). It suggests that the inferences allocated to complex images have already reached saturation.

\noindent \textbf{Initial \& local weight.} We investigate the impact of adjusting the ratio $\alpha$ between initial and local labels when updating the pseudo-labels (Eq.~\ref{eqn:weightsum}). In Fig.~\ref{fig:plot} (b), the results indicate that utilizing both global and local labels together leads to performance enhancement. This trend arises because two different label sources have complementary properties. The local label contains more information through multiple CLIP inferences of different local patches. However, since the reliability of CAM is not 100\%, it may draw an invalid patch. Conversely, the initial label contains less information, but it is always generated from valid images. By blending these two sources, we reinforce consistent predictions between local and initial labels, thereby strengthening the signal of reliable labels. Additionally, we highlight that the use of our local labels alone (\ie at weight 0) can outperforms CDUL. This confirms that our choice of local views and their labels extracts more informative learning signals than the mixture of the global view and uniformly sampled local views.

\begin{table}[!t]
    \centering{
    \caption{Per-class AP in (\%) for without and with debiasing. We focus on the seven classes from the lowest mean probability. The result showcases that our debiasing is effective for enhancing the performance of biased classes.}
    \label{tab:biasedclass}
\begin{adjustbox}{width=\linewidth}
    \begin{tabular}%{|l|l|l|l|l|}
        {
    %   |
      >{}p{0.2\linewidth}|
      >{\centering}p{0.11\linewidth}|
     >{\centering}p{0.11\linewidth}|
      >{\centering}p{0.11\linewidth}|
     >{\centering}p{0.11\linewidth}|
      >{\centering}p{0.11\linewidth}|
      >{\centering}p{0.11\linewidth}|
      >{\centering\arraybackslash}p{0.11\linewidth}
    %   |
    }
\hline

        Method & bottle & chair & table & person & plant & sofa & tv \\ \hline \hline
        %%%%%%%%%%%%%%%%%%%%%%%%%%%%%%%%%%%%%
        w/o debias & 77.0 & 77.1 & 72.7 & 87.3 & 68.8 & 73.0 & 89.5 \\ \hline
        W debias    & 77.6 & 79.0 & 74.8 & 89.2 & 72.1 & 76.6 & 92.0 \\ \hline
        
%\hline

    \end{tabular}
        \end{adjustbox}
    }
\end{table}

\noindent \textbf{Per-class mAP of biased classes.} We assess the effectiveness of our proposed method by examining the improvement in performance for specific target classes before and after CLIP debiasing. In Tab.~\ref{tab:biasedclass}, we focus on seven classes from the biased class, including \textit{bottle}, \textit{chair}, \textit{table}, \textit{person}, \textit{potted plant}, \textit{sofa}, and \textit{tv monitor} as confirmed in the Fig.~\ref{fig:bias}. Our results demonstrate consistent improvements in the performance of all these classes, ranging from 0.6\%p (\textit{bottle}) to 3.6\%p (\textit{sofa}). Moreover, we observe comparable or slightly enhanced performance for the remaining classes not shown in the table. The result affirms the effectiveness of our method across a broader spectrum. Detailed class-wise performance results for the entire dataset are provided in the supplementary material.

\subsection{Limitation}

A limitation of proposed method is that the generated local views often struggle to distinguish between co-occurring objects, an issue inherited from CAM. This challenge is especially evident in complex datasets like COCO, where similar classes--such as \textit{skies}, \textit{snowboards}, and \textit{skateboards}--frequently appear in scene with human activity. These classes often co-occur with human feet, leading classifiers to associate high activations near the feet with these classes. This phenomenon makes them potential candidates for local views. This redundant information results in noisy pseudo-labels, reducing overall performance. Examples of such failure cases are provided in the supplementary materials. Developing an improved local view proposal method to better handle noisy inferences and focusing on the target objects will be an important direction for future work.
\section{Conclusion}
In this paper, we propose a novel unsupervised multi-label classification method, Classifier-guided CLIP Distillation (CCD). We address two critical issues in CLIP: inconsistent predictions due to subtle input changes and inherent prediction bias. To tackle these issues, we devised a classifier-guided label update and CLIP debiasing. Firstly, we identified the effectiveness of acquiring local views near class-related objects and proposed a classifier-guided label update. It selects local views utilizing the Class Activation Mapping (CAM) and acquires corresponding labels from CLIP. Our approach results in allocating more local views for complex images. Secondly, we address the CLIP's bias. We investigated the biased prediction distribution of CLIP and rectified the pseudo-labels. This step notably boosts performance in certain classes and reduces the impact of bias on CLIP. Through experiments on four benchmarks, CCD newly achieved state-of-the-art performances in unsupervised multi-label classification. Remarkably, CCD achieved comparable performance to fully supervised methods on PASCAL VOC 2012 and PASCAL VOC 2007 datasets without relying on annotations. %For future work, we plan to enhance performance in datasets with numerous classes.

\noindent\textbf{Acknowledgements.} This research was supported by the Basic Science Research Program through the National Research Foundation of Korea (NRF) funded by the MSIP (RS-2025-00520207, RS-2023-00219019), KEIT grant funded by the Korean government (MOTIE) (No. 2022-0-00680, No. 2022-0-01045), the IITP grant funded by the Korean government (MSIT) (No. RS-2024-00457882, National AI Research Lab Project, No. 2021-0-02068 Artificial Intelligence Innovation Hub, RS-2019-II190075 Artificial Intelligence Graduate School Program (KAIST)), Samsung Electronics Co., Ltd (IO230508-06190-01) and SAMSUNG Research, Samsung Electronics Co., Ltd.

\clearpage
\setcounter{page}{1}
\maketitlesupplementary

\section{Detailed Analysis}

\begin{figure}[t]
\begin{center}
\includegraphics[width=\linewidth]{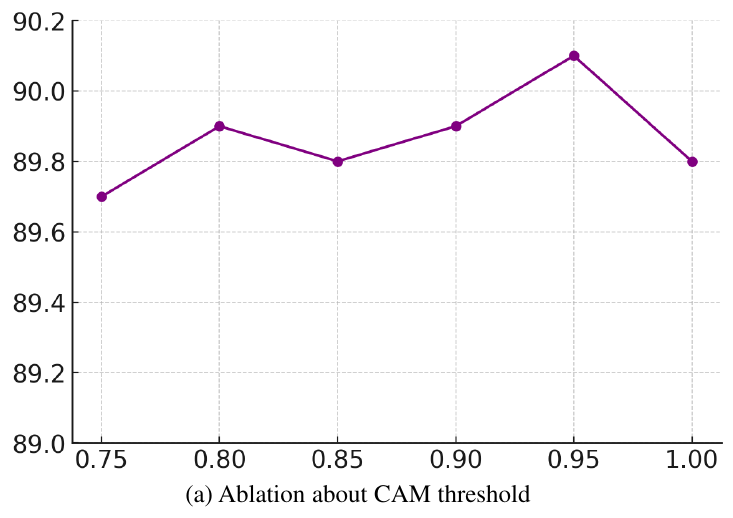}
\end{center}
\caption{Ablation study on CAM threshold. The performance remains consistent around an mAP of 90\%, with the peak near a threshold value of 0.95.}
\label{fig:sup-abl}
\end{figure}

\subsection{Ablation of CAM threshold}
Fig.~\ref{fig:sup-abl} illustrates the changes in mAP relative to variations in the CAM threshold. The peak performance is observed at a threshold value of 0.95, with consistently high performance around this point. By using CAM-based local patch acquisition, we can sample patches near around regions of interest. As the threshold decreases, performance slightly declines due to the larger local patches, which include more and more regions. This transition shifts from sampling \textit{around GT boxes} to \textit{randomly sampled boxes} which visualized in the proof-of-concept study. This ablation result aligns with the proof-of-concept study, demonstrating that sampled boxes around GT reflect the key outcomes.

\begin{figure}[t]
\begin{center}
\includegraphics[width=\linewidth]{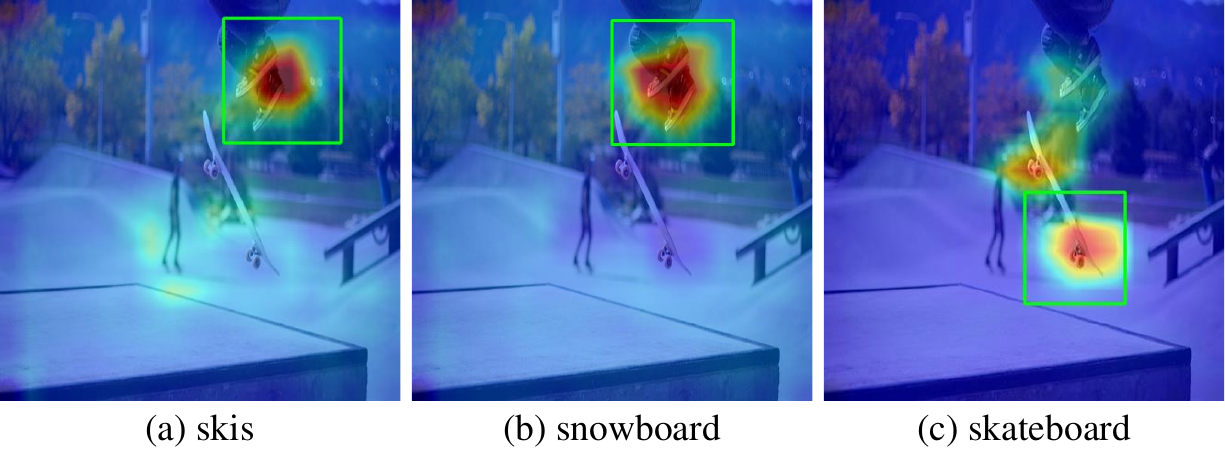}
\end{center}
\caption{CAM corresponding to different classes. The similar class co-occur with human-feet, resulting similar CAM around the feet.}
\label{fig:sup-limit}
\end{figure}

\begin{table*}[!ht]
    \normalsize
    \centering{
    \caption{AP and mAP (in \%) of unsupervised methods on PASCAL VOC 2012 dataset for all classes. The best score is in bold.}
      \label{tab:sup-per_class}
\begin{adjustbox}{width=1\textwidth}
    \begin{tabular}%{lccccccccccccccccccccc}
            {
            %|
      >{}p{0.1\textwidth}
      >{\centering}p{0.065\textwidth}
     >{\centering}p{0.065\textwidth}
      >{\centering}p{0.065\textwidth}
     >{\centering}p{0.065\textwidth}
      >{\centering}p{0.065\textwidth}
         >{\centering}p{0.065\textwidth}
     >{\centering}p{0.065\textwidth}
      >{\centering}p{0.065\textwidth}
     >{\centering}p{0.065\textwidth}
      >{\centering}p{0.065\textwidth}
          >{\centering}p{0.065\textwidth}
     >{\centering}p{0.065\textwidth}
      >{\centering}p{0.065\textwidth}
     >{\centering}p{0.065\textwidth}
      >{\centering}p{0.065\textwidth}
         >{\centering}p{0.065\textwidth}
     >{\centering}p{0.065\textwidth}
      >{\centering}p{0.065\textwidth}
     >{\centering}p{0.065\textwidth}
      >{\centering}p{0.065\textwidth}  
      >{\centering\arraybackslash}p{0.065\textwidth}
    %   |
    }
    \hline
    
         Methods& aero & bicycle & bird & boat & bottle & bus & car & cat & chair & cow & table & dog & horse & bike & person & plant & sheep & sofa & train & tv & mAP \\ \hline

        CDUL & 99.0 &92.7 &\textbf{97.7} &91.8 &72.5 &95.4 &84.7 &98.6 &76.4 &91.9 &73.2 &97.1 & 92.0 & 94.1 &\textbf{93.0} &67.5 & 94.2 &74.2 & 97.7 &89.0 &88.6\\

        w/o Debias & 98.4 & 92.8 & 97.6 & 91.5 & 77.0 & 96.0 & 83.9 & 98.7 & 77.1 & 93.4 &  72.7 & 96.5 & 96.1 & 93.7 & 87.3 & 68.8 & 94.0 & 73.0 & 98.3 & 89.5 & 88.8\\
         
        CCD (ours)  & \textbf{99.1} &\textbf{93.6} &97.6 &\textbf{92.4} &\textbf{77.6} &\textbf{96.0} &\textbf{86.1} &\textbf{99.0} &\textbf{79.0} &\textbf{95.2} &\textbf{74.8} &\textbf{97.7} & \textbf{96.0} & \textbf{95.4} &89.2 &\textbf{72.1} & \textbf{95.3} &\textbf{76.6} & \textbf{98.1} &\textbf{92.0} & \textbf{90.1}\\ \hline
    \end{tabular}
    \end{adjustbox}
    }
\end{table*}

\begin{table*}[ht]
  \centering
  \caption{AP and mAP (in \%) of unsupervised multi-label classification on MS COCO 2014 dataset for all classes.}
  \normalsize
  \begin{adjustbox}{max width=\textwidth}
    \begin{tabular}{lc@{\hskip 0.1in}c@{\hskip 0.1in}|@{\hskip 0.1in}lc@{\hskip 0.1in}c@{\hskip 0.1in}|@{\hskip 0.1in}lc@{\hskip 0.1in}c@{\hskip 0.1in}|@{\hskip 0.1in}lc@{\hskip 0.1in}c@{\hskip 0.1in}|@{\hskip 0.1in}lc@{\hskip 0.1in}c}
    \hline
    Class & w/o DB & Ours & Class & w/o DB & Ours & Class & w/o DB & Ours & Class & w/o DB & Ours & Class & w/o DB & Ours \\
    \hline
    \hline
    
    person      & 81.7 & 82.0 &  horse         & 90.4 & 91.7 &  baseball bat            & 89.4 & 89.0 &  carrot       & 48.7 & 64.1 &  microwave      & 63.4 & 65.8 \\
    bicycle          & 62.9 & 69.8 &  sheep        & 94.7 & 94.3 &  baseball glove    & 83.0 & 86.9 &  hot dog         & 67.3 & 69.5 &  oven       & 69.3 & 67.6 \\
    car         & 59.2 & 61.3 &  cow        & 87.4 & 89.4 &  skateboard  & 95.8  & 96.2  &  pizza        & 92.1 & 93.6 &  toaster            & 1.9 & 5.7 \\
    motorcycle             & 86.3 & 88.8 &  elephant         & 97.4 & 98.0 &  surfboard      & 93.0 & 94.2 &  donut          & 76.1 & 79.2 &  sink         & 83.8 & 83.4 \\
    airplane      & 95.5 & 96.8 &  bear     & 95.8 & 94.8 &  tennis racket       & 98.6 & 98.6 &  cake          & 77.4 & 75.5 &  refrigerator            & 68.4 & 70.1 \\
    bus        & 81.3 & 81.7 &  zebra         & 97.8 & 99.0 &  bottle   & 40.1 & 45.0 &  chair           & 50.3 & 54.2 &  book    & 28.2 & 26.0 \\
    train             & 94.7 & 95.5 &  giraffe        & 97.9 & 98.9 &  wine glass    & 56.5 & 64.1 &  couch          & 68.3 & 71.7 &  clock            & 74.6 & 76.2 \\
    truck           & 55.9 & 54.8 &  backpack      & 20.7 & 32.9 &  cup      & 36.2 & 38.0 &  potted plant          & 37.5 & 45.3 &  vase           & 65.8 & 67.4 \\
    boat           & 81.9 & 85.9 &  umbrella     & 74.8 & 78.7 &  fork             & 42.1 & 53.3 &  bed   & 79.5 & 79.6 &  scissors            & 47.6 & 49.8 \\
    traffic light            & 74.2 & 76.9 &  handbag     & 20.6 & 25.7 &  knife            & 32.8 & 38.6 &  dining table            & 48.8 & 48.1 &  teddy bear        & 77.2 & 84.2 \\
    fire hydrant   & 78.1 & 79.2 &  tie      & 68.5  & 70.3  &  spoon           & 33.0 & 40.2 &  toilet   & 95.6 & 96.3 &  hair drier      & 27.7 & 22.2 \\
    stop sign    & 72.3 & 73.1 &  suitcase         & 62.8 & 65.6 &  bowl           & 34.7  & 38.5  &  tv         & 74.6 & 77.4 &  toothbrush      & 57.3 & 59.5 \\
    parking meter       & 60.8 & 63.7 &  frisbee     & 88.1 & 90.9 &  banana    & 74.5 & 77.6 &  laptop             & 80.5 & 84.2 &  \multicolumn{1}{c}{}  & \multicolumn{1}{c}{} & \multicolumn{1}{c}{} \\
    bench   & 52.0 & 53.6 &  skis      & 91.6 & 91.3 &  apple          & 45.9 & 46.9 &  mouse         & 48.1 & 45.9 &  \multicolumn{1}{c}{} & \multicolumn{1}{c}{} & \multicolumn{1}{c}{}\\
    \cline{13-15}
    bird           & 69.4 & 75.5 &  snowboard         & 72.8  & 75.9 & sandwich            & 70.0 & 73.9 &  remote          & 49.2 & 58.3 &  \multicolumn{1}{l}{\multirow{4}[0]{*}{mean}} & \multicolumn{1}{c}{\multirow{4}[0]{*}{67.7}} & \multicolumn{1}{c}{\multirow{4}[0]{*}{70.3}}\\
    cat            & 91.4 & 93.6 &  sports ball    & 33.9 & 30.1 & orange        & 52.6 & 62.1 &  keyboard         & 75.3 & 77.1 &  \multicolumn{1}{c}{} & \multicolumn{1}{c}{} & \multicolumn{1}{c}{} \\
    dog             & 76.8 & 81.9 &  kite  & 95.0 & 93.9 &  broccoli          & 89.7 & 90.3 &  cell phone      & 53.0 & 56.7 &  \multicolumn{1}{c}{} & \multicolumn{1}{c}{} & \multicolumn{1}{c}{} \\
    \hline
    \end{tabular}%
    \end{adjustbox}%
  
  \label{tab:sup_per-class_coco}%
\end{table*}%

\subsection{Per-class Performance}
\begin{figure*}[h]
\begin{center}
\includegraphics[width=8cm]{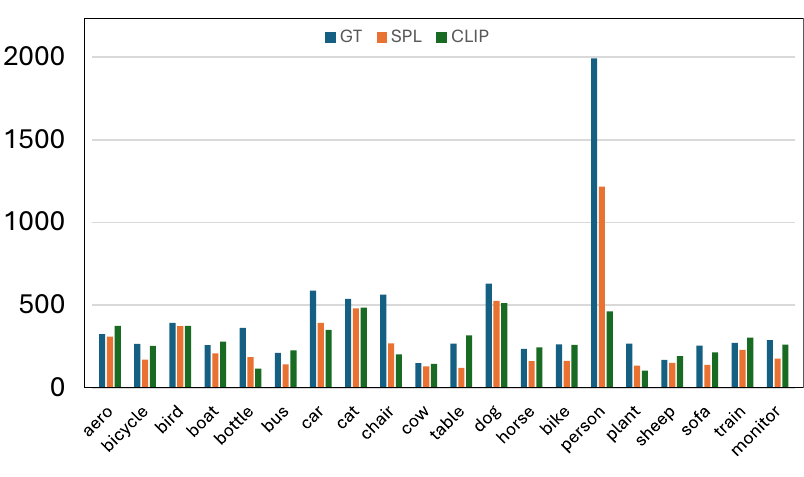}
\end{center}
\caption{The label distribution of GT, Single Positive Label, CLIP generated at PASCAL VOC 2012. For the CLIP generated label, we count top-1 as predicted. Single Positive setting reflects the label distribution of GT labels, which cannot reflect the real behavior of labelers.}
\label{fig:sup-label}
\end{figure*}

Tab.~\ref{tab:sup-per_class} shows the per-class average precision on the PASCAL VOC 2012 dataset. We investigate the impact of the CLIP debiasing and observe that the bias correction is effective for boosting the performance of biased classes. Furthermore, the performance of unbiased classes also improves alongside biased classes. This is attributed to correcting mispredicted biased classes, which would otherwise introduce noise when predicted as other classes. However, the amount of improved performance differs depend on the classes. Regarding the performance of \textit{bottle}, \textit{sofa}, and \textit{tv}, the low performance for \textit{bottle} can be attributed to its greater CLIP bias (relatively lower probabilities). In addition, the fact that \textit{bottle} frequently appears but its detection rate remains low as illustrated in Fig.~\ref{fig:sup-label}. On the other hand, \textit{sofa} and \textit{tv} classes captured as top-1 are similar to GT quantities.

Compared to CDUL, our CCD outperforms all classes except \textit{bird} and \textit{person}. This might be due to the gradient-alignment network training method from CDUL, which updates the labels during the whole training process with training loss. Since the person label presents the most frequently in the PASCAL VOC dataset (shown in Fig.~\ref{fig:sup-label}), the training process is likely to update the person label to ``Positive''.

Tab.~\ref{tab:sup_per-class_coco} shows the per-class average precision on the MS COCO dataset. Due to undisclosed hyperparameters, we cannot reproduce the CDUL. Thus, we only compare the results of CCD and CCD without debiasing. Our proposed CLIP debiasing successfully boosts the performance of biased classes.

\subsection{Bounding Box with Class Activation Mapping}

Our method generates CAM bounding boxes for classes that exceed a certain threshold, allowing for multiple bounding boxes to be obtained around objects. Figure~\ref{fig:sup-cam} shows sample bounding boxes extracted from an image in the PASCAL VOC 2012 dataset. The first row represents the best-case scenario, where multiple bounding boxes accurately surround the target classes. The second row illustrates a mixed scenario, with half of the boxes correctly identifying the target classes and the other half missing them. The third row represents a failure case, where only a single bounding box is close to the target object. These bounding boxes are then cropped and passed through CLIP to generate local labels. %two bounding boxes are generated near the sofa, one around the person, and one in the vicinity of the TV monitor. Due to potential noise in the classifier predictions, only the region of interest information is utilized.

\section{Dataset}

\subsection{Probability Distribution}
\begin{figure*}[ht]
\begin{center}
\includegraphics[width=12cm]{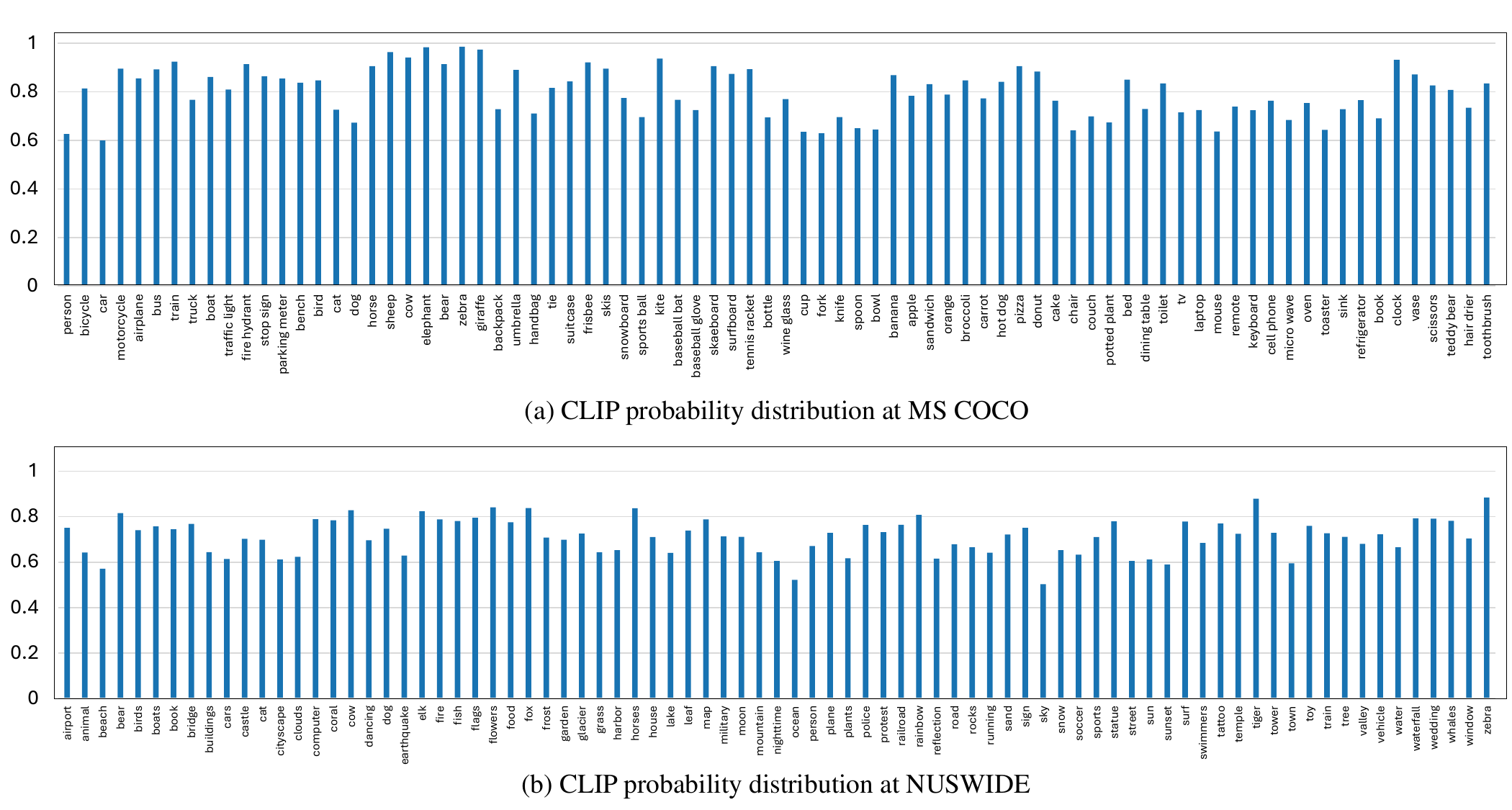}
\end{center}
\caption{CLIP probability distribution for other datasets: (a) The mean class-wise probability of MS COCO. (b) The mean class-wise probability of NUSWIDE. The probability distribution showcases the presence of CLIP bias in other datasets.}
\label{fig:sup-distribution}
\end{figure*}

Fig.~\ref{fig:sup-distribution} illustrates the CLIP probability distribution for MS COCO and NUSWIDE datasets. It is evident that the mean probability of NUSWIDE is comparatively lower than that of MS COCO. This discrepancy likely contributes to the lower classifier performance observed in NUSWIDE compared to MS COCO. Additionally, we note that the same ``class name'' exhibits similar low mean probability across both datasets (\eg the probability of \textit{person} is 0.63 for COCO, 0.67 for NUSWIDE). This observation underscores the presence of bias inherent to the CLIP model and text embedding, irrespective of the dataset.

\subsection{Label Distribution}

Fig.~\ref{fig:sup-label} shows the label distribution for PASCAL VOC 2012. We can observe that the label distribution of GT and Single Positive Label (SPL) is similar. However, the CLIP generated label shows does not reflect GT label's distribution. In particular, the number of \textit{person} prediction is relatively lower. This result combine with the CLIP bias, making the prediction of biased class harder. This results support that CLIP shows biased prediction. 

\section{Limitation}
\begin{figure*}[h]
\begin{center}
\includegraphics[width=12cm]{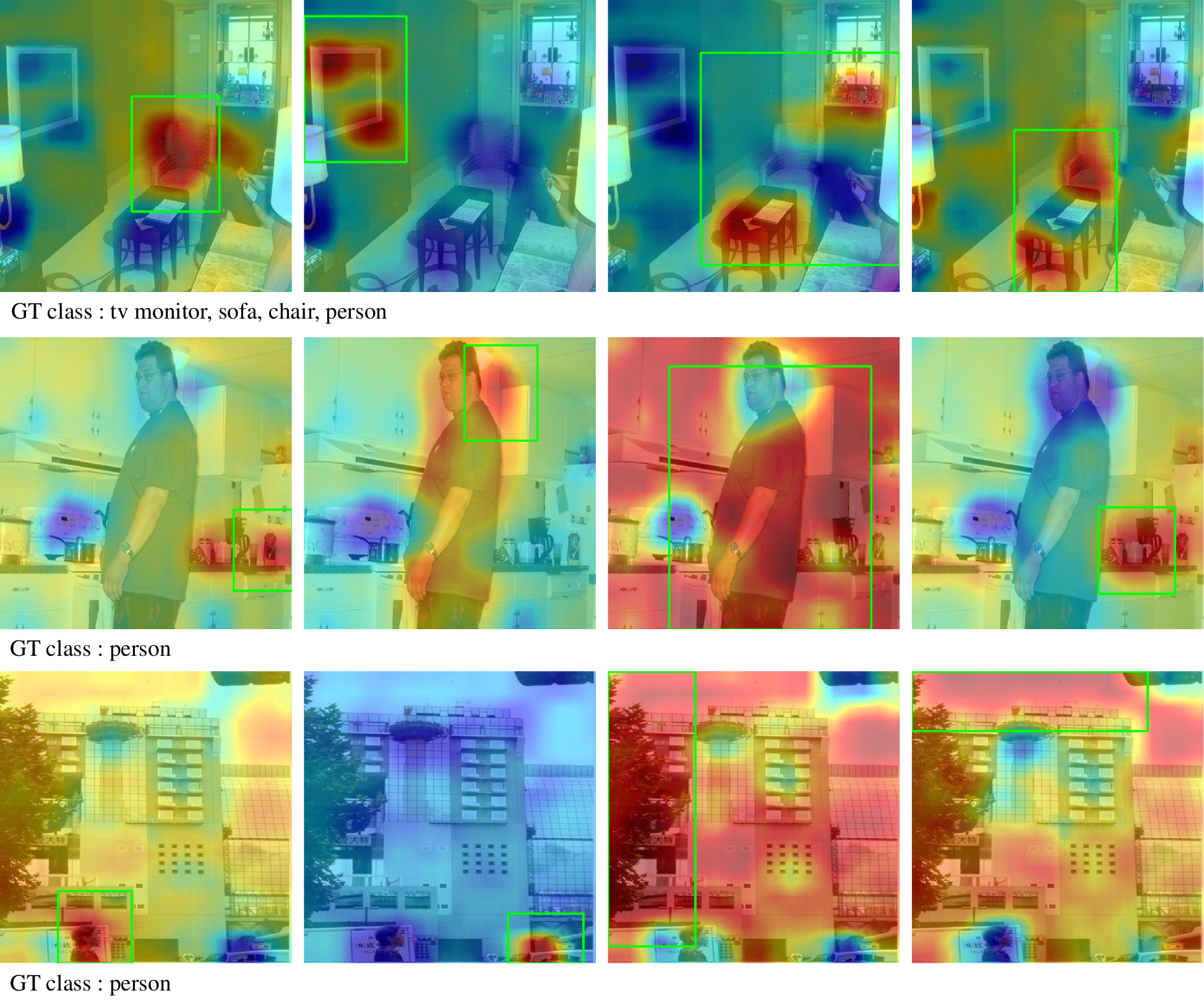}
\end{center}
\caption{The sample bounding box with Class Activation Mapping. The first row shows optimal case, where every boxes capture gt-related objects. The second row illustrates mixed case, where half of the boxes captures \textit{person} class. The third row depict failure case, where only one box contains \textit{person} class. These multiple local inferences around objects help make better pseudo-labels.}
\label{fig:sup-cam}
\end{figure*}

A limitation of proposed method is that the generated local views often struggle to distinguish between co-occurring objects, an issue inherited from CAM. In Fig.~\ref{fig:sup-limit}, the activation map of \textit{skies} and \textit{snowboard} focus on the human feet. This phenomenon makes them potential candidates for local views. This redundant information results in noisy pseudo-labels, reducing overall performance. Moreover, in Fig.~\ref{fig:sup-cam}, the second row shows the \textit{bottle} (the 1st image) or \textit{dinning table} (the 4th image) generates false local patches. Developing an improved local view proposal method to better handle noisy inferences and focusing on the target objects will be an important direction for future work.

In addition, it is evident from Tab.1 in main text that all unsupervised multi-label classification methods exhibit relatively degraded performance on MS COCO and NUSWIDE datasets. Both datasets share the characteristic of having a larger number of classes (80 and 81, respectively) compared to the PASCAL VOC datasets (20 classes). This performance degradation can be attributed to the simplistic prompt, ``a photo of the [class name],'' which might not accurately represent the target classes. For instance, the performance of initial labels of the training sets for PASCAL VOC 2012, MS COCO, and NUSWIDE is 85.3\%, 65.4\%, and 41.2\%, respectively. 

This suggests that CLIP predictions already show varying performance across datasets with different class counts (Although MS COCO and NUSWIDE have similar class counts, the inclusion of ``none images'' in NUSWIDE exacerbates the degradation of CLIP performance). To address this limitation, manipulating the text embeddings of each dataset may be necessary. 

For ``none images,'' the pseudo-labels should be set to zero. To address this, we suggest filtering out the probability of ``none images,'' as a possible solution. Specifically, we can identify irregular and noisy probability patterns from those images and then apply thresholding to filter them out. %This will be a promising future work. 

% \section{Rationale}
% \label{sec:rationale}
% % 
% Having the supplementary compiled together with the main paper means that:
% % 
% \begin{itemize}
% \item The supplementary can back-reference sections of the main paper, for example, we can refer to \cref{sec:intro};
% \item The main paper can forward reference sub-sections within the supplementary explicitly (e.g. referring to a particular experiment); 
% \item When submitted to arXiv, the supplementary will already included at the end of the paper.
% \end{itemize}
% % 
% To split the supplementary pages from the main paper, you can use \href{https://support.apple.com/en-ca/guide/preview/prvw11793/mac#:~:text=Delete%20a%20page%20from%20a,or%20choose%20Edit%20%3E%20Delete).}{Preview (on macOS)}, \href{https://www.adobe.com/acrobat/how-to/delete-pages-from-pdf.html#:~:text=Choose%20%E2%80%9CTools%E2%80%9D%20%3E%20%E2%80%9COrganize,or%20pages%20from%20the%20file.}{Adobe Acrobat} (on all OSs), as well as \href{https://superuser.com/questions/517986/is-it-possible-to-delete-some-pages-of-a-pdf-document}{command line tools}.

\end{document}